# Tractability of Theory Patching

**Shlomo Argamon-Engelson**                    ARGAMON@CS.BIU.AC.IL
**Moshe Koppel**                               KOPPEL@CS.BIU.AC.IL
*Department of Mathematics and Computer Science*
*Bar-Ilan University*
*52900 Ramat Gan, Israel*

## Abstract

In this paper we consider the problem of *theory patching*, in which we are given a domain theory, some of whose components are indicated to be possibly flawed, and a set of labeled training examples for the domain concept. The theory patching problem is to revise only the indicated components of the theory, such that the resulting theory correctly classifies all the training examples. Theory patching is thus a type of theory revision in which revisions are made to individual components of the theory. Our concern in this paper is to determine for which classes of logical domain theories the theory patching problem is tractable. We consider both propositional and first-order domain theories, and show that the theory patching problem is equivalent to that of determining what information contained in a theory is *stable* regardless of what revisions might be performed to the theory. We show that determining stability is tractable if the input theory satisfies two conditions: that revisions to each theory component have monotonic effects on the classification of examples, and that theory components act independently in the classification of examples in the theory. We also show how the concepts introduced can be used to determine the soundness and completeness of particular theory patching algorithms.

## 1. Introduction

In this paper we consider the problem of *theory patching*, in which we are given a domain theory, some of whose components are indicated to be possibly flawed, and a set of labeled training examples for the domain concept. The theory patching problem is to revise only the indicated components of the theory, such that the resulting theory correctly classifies all the training examples. Our concern in this paper is to determine for which classes of logical domain theories the theory patching problem is tractable.

Theory patching can be thought of as a type of theory revision in which revisions are made to individual components of the theory. Many such algorithms have been investigated in the literature on theory revision, both in machine learning (Koppel, Feldman, & Segre, 1994; Ourston & Mooney, 1994; Saitta, Botta, & Neri, 1993; Wogulis, 1991) and in inductive logic programming (ILP) (Adé, Malfait, & DeRaedt, 1994; DeRaedt, 1992; Wrobel, 1994, 1995). Previous work in theory revision has primarily been concerned with the problem of finding the optimal revision to a theory for a given set of training examples, according to some preference bias. For example, a theory revision algorithm may prefer to make the fewest number of revisions necessary to satisfy all of the training examples. We, however, are not concerned with finding an optimal set of revisions, but rather with finding *any* satisfying set of revisions, under the constraint that some theory components may not be revised.

Note that if all components in a theory are revisable, the theory patching problem is trivially solvable, since the theory can simply be discarded and replaced by some theory which satisfies all the examples (if such a theory exists). Hence the problem is interesting only because of the inductive bias consisting of restricting revision to a subset of the input theory's components. Such a restriction is often used as an initial restriction on the revision process (Koppel et al., 1994; Saitta et al., 1993; Weber & Tausend, 1994), but also can arise during the process of revision when an algorithm performs an unretractable revision. Thus the tools developed in this paper for analyzing the theory patching problem may be applied to many theory revision algorithms extant in the literature.

We consider both propositional and first-order domain theories, and show that the theory patching problem is equivalent to that of determining what information contained in a theory is *stable* regardless of what revisions might be performed to the theory. In particular, if some example is stably misclassified (*i.e.*, it remains misclassified regardless of how the theory may be revised), then the theory cannot be patched. We will show that although determining stability is not tractable in general, it is tractable if the input theory satisfies two conditions, which will be discussed in detail in the paper. The first condition concerns the monotonicity of the theory under possible revisions, while the second condition concerns the independence of theory components in the classification of examples in the theory.

In the next section, we treat the tractability of theory patching for propositional theories, showing that the problem is intractable in general, but that a polynomial-time algorithm exists for a large subclass of propositional theories. Then in Section 3 we discuss what additional restrictions suffice to ensure tractability of patching for first-order domain-theories.

## 2. Propositional Theory Patching

In this section, we address the problem of patching propositional domain theories. We first define the problem precisely, and then show that the problem is NP-hard in general. Then, in Section 2.3 we provide an algorithm for reducing the patching problem to the stability problem. In Section 2.4 we show for what class of theories the stability problem is tractable, and give an algorithm for solving stability for such theories. We conclude this section by showing how the concepts introduced can be used to determine the soundness and completeness of particular theory patching algorithms.

### 2.1 Defining the problem

#### 2.1.1 Theories

In this section, we consider cycle-free propositional definite-clause theories with a single root proposition, allowing negation as failure in clause antecedents. Each clause $p \leftarrow l_1 \land \cdots \land l_n$ consists of a **head**, $p$, and **body**, $l_1 \land \cdots \land l_n$. For a given theory $\Gamma$, we stipulate as given a set of **primitive propositions**, such that no primitive proposition may appear in the head of a clause in the theory. Primitive propositions serve the function of 'observable facts' in the theory.

| CUP | ← | UPRIGHT ∧ LIFTABLE ∧ OPEN |
| UPRIGHT | ← | HAS-BOTTOM |
| LIFTABLE | ← | GRASPABLE ∧ LIGHT-WEIGHT |
| OPEN | ← | HAS-CONCAVITY ∧ UPWARD-CONCAVITY |
| OPEN | ← | HAS-STRAW |
| GRASPABLE | ← | HAS-HANDLE |
| GRASPABLE | ← | SMALL ∧ CERAMIC ∧ DRY |

Figure 1: The CUP theory (adapted from Winston et al. (1983)).

An **example** $E$ for a propositional theory $\Gamma$ with root $r$ is a truth assignment to all the primitive propositions in $\Gamma$. We say that $\Gamma$ **covers** $E$ if $\Gamma \cup E \vdash r$ (under standard resolution with negation as failure), otherwise $E$ is termed **uncovered** by $\Gamma$; let $\Gamma(E) = T$ if $E$ is covered by $\Gamma$, and $\Gamma(E) = F$ otherwise. Training data is given in the form of labeled examples, where a **labeled example** is a pair $\langle E, l \rangle$ consisting of an example $E$ and a truth assignment $l \in \{T, F\}$. If $l = T$, the labeled example is termed **positive**, otherwise it is termed **negative**. The goal of theory patching is to revise a given theory so that it covers all given positive examples and does not cover any given negative example.

> **Example.** Consider the "CUP theory" given in Figure 1, which classifies objects as cups or non-cups. The head of the first clause, for example, is "CUP" and its body is the expression "UPRIGHT ∧ LIFTABLE ∧ OPEN". The primitive propositions of the theory are HAS-BOTTOM, LIGHT-WEIGHT, HAS-CONCAVITY, UPWARD-CONCAVITY, HAS-STRAW, HAS-HANDLE, SMALL, CERAMIC, and DRY.
>
> Now consider the labeled example $\langle E_1, T \rangle$, where $E_1 = \{$HAS-BOTTOM, LIGHT-WEIGHT, HAS-CONCAVITY, UPWARD-CONCAVITY, SMALL$\}$. Although the labeled example is positive, $E_1$ is not covered by $\Gamma$ (*i.e.*, $\Gamma(E_1) = F$), and so the theory needs to be revised.

In theory patching we are interested in considering revisions to individual components of a theory. We thus define the **components** of a theory $\Gamma$ to be its propositions, clauses, and the literals in the antecedent of each clause (where separate appearances of the same literal are considered different components). We distinguish between a proposition itself and each of its appearances in the theory's clause antecedents since in the course of revision we may wish either to change the definition of a proposition or to remove a given appearance of the proposition without changing its definition[1].

---

1. The corresponding distinction for a clause with a conjunction of propositions in its head would be to either revise the body of the clause or to delete one of the propositions in its head. Since we only allow a single head proposition for each clause, this issue is not relevant here.

2.1.2 Revisions

Now we restrict which components of a theory may be revised, by defining:

**Definition 1** *A **patchable propositional theory** is a pair $\langle \Gamma, \Omega \rangle$ consisting of a propositional domain theory $\Gamma$ and a set $\Omega$ of components of $\Gamma$. We call $\Omega$ the **open components** of $\langle \Gamma, \Omega \rangle$, terming all components of $\Gamma$ not in $\Omega$ **closed**.*

The only components that may be revised in a patchable theory are the open components; the closed components must remain as is. Intuitively, the open components are those which we believe might be flawed, while the closed components are those which we assume are correct as is. How this information might be obtained is beyond the scope of this paper; we simply assume it as given. For example, a set of open components could be provided by the domain expert who created the theory. Closing some components of the theory amounts to an inductive bias on the learning problem, removing some possible theories (those not obtainable by revisions to open components) from consideration as hypotheses.

We now consider revisions to a propositional domain theory. Roughly speaking, we allow deletion and addition of components, but we will need to be precise in order to ensure that revisions are local to a component. We first define, therefore, what exactly constitutes a revision to a *specific component* as opposed to a global revision to the theory. We will use the notion of a **simple clause**, whose body contains literals only for primitive propositions.

- We regard deletion of a component to be a revision to that component. Thus we can revise a clause or an antecedent literal by simply deleting it from the theory. Note that we can think of deleting an antecedent literal as treating the literal as if it were always true. Hence, we naturally extend the definition to propositions by defining deletion of a non-primitive proposition $p$ to mean the addition of a clause for $p$ that renders $p$ always true. (To be precise, this has the effect of deleting every antecedent literal $p$ and deleting all clauses containing the antecedent literal $\neg p$.) We denote by $\Gamma \backslash \{c\}$ the theory resulting from the deletion of component $c$ from $\Gamma$.

- We may also revise a clause in $\Gamma$ by adding a new positive or negative literal for a new proposition, $q \notin \Gamma$, to the body of the clause, and adding a set of new simple clauses defining $q$ to $\Gamma$. (The restriction to simple clauses ensures that the effects of the revision are local to the clause under revision.)

- Similarly, we may revise a proposition $p$ by adding a new clause $C$ with head $p$ to $\Gamma$, such that the propositions appearing in $C$'s body do not appear in $\Gamma$, as well as a set of new simple clauses defining each of those new propositions. (As above, the restriction to simple clauses ensures that the effects of the revision are local to the clause under revision.)

**Example.** Consider a revision to the CUP theory which (i) adds the antecedent literal "¬ABSORB" to the body of the first GRASPABLE clause (changing the clause to read "GRASPABLE←HAS-HANDLE ∧ ¬ABSORB"), and (ii) adds clauses to the theory defining the new proposition:

- ABSORB←CERAMIC ∧ HAS-BOTTOM

- ABSORB←HAS-CONCAVITY ∧ HAS-BOTTOM

This revision (call it $r_1$) is considered a revision to the first GRASPABLE clause in the theory. On the other hand, if the revision included, say, addition of the clause "ABSORB←SMALL ∧ OPEN', it would *not* constitute a revision to the GRASPABLE clause, since the new clause is not simple (OPEN is a non-primitive proposition in the theory).

Similarly, consider the revision $r_2$, consisting of adding the clauses:

- LIFTABLE←FITS-IN-HAND

- FITS-IN-HAND←SMALL ∧ ¬HAS-STRAW

This revision is considered a revision to the proposition LIFTABLE, since the clause added for the new proposition is simple. If, however, the revision included, say, addition of the clause "FITS-IN-HAND←SMALL ∧ GRASPABLE" it would not constitute a revision to that proposition, since the added clause is not simple (and hence the revision affects more than one component of the theory).

Note that in the above example, after we apply revision $r_1$, the first GRASPABLE clause cannot be used to prove any example for which ¬ABSORB is false (*i.e.*, for which HAS-BOTTOM and either CERAMIC or HAS-CONCAVITY are true). If for some such example there are no proofs in the original theory which do not use the clause being revised, then the example will not be covered by the revised theory, even if it was covered by the original theory. Thus we can say that the first GRASPABLE clause is 'disabled' by the revision $r_1$ for such examples. Note that the revision does not change the classification of examples for which ¬ABSORB is true, and hence the clause is not disabled for those examples.

Similarly, after the application of $r_2$, the proposition LIFTABLE cannot prevent a proof of any example for which FITS-IN-HAND is true (*i.e.*, for which SMALL is true and HAS-STRAW is false). If such an example was uncovered in the original theory only because LIFTABLE was false, then in the revised theory the example becomes covered. That is, we can say that the proposition LIFTABLE becomes 'disabled' for such examples, in the sense that, for those examples, the revised theory acts just like a theory with the proposition deleted. For those examples for which FITS-IN-HAND is false, however, the proposition is not disabled by the revision.

More generally, in the case of adding a literal to a clause we can say that the clause is effectively disabled for all examples for which the added literal is false. Hence, in the case of a revision to a clause, the set of examples for which the new antecedent literal is false is called the **disabling set** of the revision. Similarly, in the case of a revision adding a new clause for a proposition we can say that the proposition is effectively disabled for all examples for which the body of the new clause is satisfied. Hence, in the case of revision to a proposition, the set of examples for which the body of the new clause is satisfied is called the **disabling set** of the revision.

> **Example.** The disabling set of the revision $r_1$ described above, adding the literal ¬ABSORB to the GRASPABLE clause, is the set of all examples for which ¬ABSORB is false, *i.e.*, those examples for which HAS-BOTTOM and either CERAMIC or HAS-CONCAVITY are true. Similarly, the disabling set of the revision $r_2$ described above, adding the clause LIFTABLE←FITS-IN-HAND for the LIFTABLE proposition, is the set of all examples for which FITS-IN-HAND is true, *i.e.*, those examples for which SMALL is true and HAS-STRAW is false.

Note that any two revisions to a component that disable the same set of examples are identical in their effects, regardless of their syntactic form. Thus, for our purposes a revision to a component is defined by its disabling set. For example, any revision that disables a component for all examples is called 'deletion', while any revision that does not disable a component for any example is called the 'null revision'.

We will consider the general problem setting in which each component in a theory may have a different set of revisions that may be applied to that component. That is, some revisions may be permitted at certain components but not at other components. We stipulate that the set of **permitted** revisions to a patchable theory $\langle \Gamma, \Omega \rangle$ is given by a **revision function** $\rho$ from each component in $\Omega$ to the set of revisions that may be applied to that component; for a theory component $c$, $\rho(c)$ is the set of revisions that may be applied to $c$.

The most general possible set of revisions is one in which any open proposition or clause in the theory may be disabled for any set of examples. Therefore, we say that a revision function $\rho$ is **unrestricted** if for each open proposition or clause component $c \in \Omega$ and any set of examples $\mathcal{E}$, there exists $r \in \rho(c)$ such that $\mathcal{E}$ is the disabling set of $r$.

### 2.1.3 PATCHING

Given a patchable theory and a set of examples, we wish to find some theory that is consistent with the examples, such that the theory is obtainable from the input theory by sequences of permitted revisions. More formally:

**Definition 2** *Given a patchable theory $\langle \Gamma, \Omega \rangle$ and revision function $\rho$, a theory $\Gamma'$ is $\rho$-**obtainable** from $\Gamma$ if, for some integer $n$, there exists a sequence of $n$ revisions $\mathcal{R} = \{r_i\}$ to respective components $\{c_i\} \subset \Omega$, where $r_i \in \rho(c_i)$, such that $\Gamma' = r_n(c_n, r_{n-1}(c_{n-1}, \ldots, r_1(c_1, \Gamma)))$.*

For convenience, if $\rho$ is unrestricted, we drop the $\rho$ and simply say that $\Gamma'$ is **obtainable** from $\Gamma$.

The patching problem, then, is to revise open components so that all of the given positive and none of the given negative examples are covered by the new theory:

**Definition 3** *The **propositional theory patching problem** $PPATCH(\Gamma, \Omega, \rho, \mathcal{E})$ is:*

**Given:** *a patchable propositional theory $\langle \Gamma, \Omega \rangle$, revision function $\rho$, and set $\mathcal{E}$ of labeled examples for $\Gamma$,*

**Find:** *a theory $\Gamma'$ which is $\rho$-obtainable from $\Gamma$ such that for all $\langle E, l \rangle \in \mathcal{E}$, $\Gamma'(E) = l$, if such a theory exists; otherwise, return FAIL.*

If such a $\rho$-obtainable theory exists, we say that $\langle \Gamma, \Omega \rangle$ is **repairable** under $\rho$ for $\mathcal{E}$.

When $\rho$ is an unrestricted revision function, we drop the $\rho$ and write $PPATCH(\Gamma, \Omega, \mathcal{E})$. In this case, the only restriction on possible revisions is given by the set of open components $\Omega$.

Note that in the case of unrestricted revisions a single revision to each open component is in principle sufficient, since a single revision can disable a component for an arbitrary subset of examples. Moreover, the order of revisions is irrelevant. Unrestricted revisions are assumed in many theory revision systems in the machine learning literature (Koppel et al., 1994; Ourston & Mooney, 1994; Richards & Mooney, 1991), although in the ILP literature, various restrictions on the revision function are assumed (see Wrobel (1995) and other works referenced there). In this paper the hardness (Section 2.2) and soundness results (Section 2.5) hold for a wide range of restricted revision functions, while the tractability results (Section 2.3) are shown for the case of unrestricted revision functions.

> **Example.** Suppose we are given a set of open components $\Omega$ for the CUP theory consisting of the first clause for GRASPABLE and the antecedent literal CERAMIC[2] in the last clause for GRASPABLE (depicted in Figure 2 by underlining). Let us also assume as given an unrestricted revision function. This represents the situation in which the expert who gave us the theory is somewhat unsure of her definition of GRASPABLE, and so indicates that its first clause may be too permissive, and also that the antecedent literal CERAMIC in the last clause for GRASPABLE may not be necessary. That is, the theory as given is the best one the expert can currently give, but she allows that some parts of the theory may be mistaken. For example, the fact that the antecedent literal HAS-HANDLE is not marked as possibly flawed means that if the first GRASPABLE clause is included, that antecedent literal must appear; the clause might be deleted entirely, or other antecedent literals may be added to it, but the HAS-HANDLE antecedent literal may not be deleted.
>
> Consider now the example set $\mathcal{E}$ consisting of the examples:

$E_1$: {HAS-BOTTOM, LIGHT-WEIGHT, HAS-CONCAVITY, UPWARD-CONCAVITY, SMALL, DRY}, labeled as positive,

$E_2$: {HAS-BOTTOM, LIGHT-WEIGHT, HAS-STRAW, HAS-HANDLE, HAS-CONCAVITY, UPWARD-CONCAVITY, SMALL, CERAMIC}, labeled as negative, and

---

2. Note that it is the *antecedent literal* CERAMIC and not the proposition that is open here. The primitive proposition CERAMIC is not revisable.

| | | |
|---|---|---|
| CUP | ← | UPRIGHT ∧ LIFTABLE ∧ OPEN |
| UPRIGHT | ← | HAS-BOTTOM |
| LIFTABLE | ← | GRASPABLE ∧ LIGHT-WEIGHT |
| OPEN | ← | HAS-CONCAVITY ∧ UPWARD-CONCAVITY |
| OPEN | ← | HAS-STRAW |
| GRASPABLE | ← | HAS-HANDLE |
| GRASPABLE | ← | SMALL ∧ CERAMIC ∧ DRY |

Figure 2: The patchable CUP theory with open components underlined.

$E_3$: {HAS-BOTTOM, LIGHT-WEIGHT, HAS-STRAW, HAS-HANDLE, HAS-CONCAVITY, SMALL}, labeled as positive, and

$E_4$: {HAS-BOTTOM, LIGHT-WEIGHT, HAS-STRAW, HAS-HANDLE, HAS-CONCAVITY, UPWARD-CONCAVITY, SMALL}, labeled as positive.

The theory $\Gamma$ does not cover the example $E_1$. However, $\Gamma$ may be revised to cover $E_1$ by the deletion of the open antecedent literal CERAMIC, without harming the desired classifications of other examples in $\mathcal{E}$. Similarly, although $\Gamma$ incorrectly covers the negative example $E_2$, it can be revised to not cover the example by the deletion of the first GRASPABLE clause; however, such a revision would cause examples $E_3$ and $E_4$ to be incorrectly classified. A proper revision to the clause, that uncovers $E_2$ while keeping $E_3$ and $E_4$ covered, would be to add the antecedent literal ¬X to it, along with adding the clause "X←SMALL ∧ HAS-STRAW". The new literal is true for both $E_3$ and $E_4$, and so only $E_2$ is made uncovered by the revision. In other words, $E_2$ is in the disabling set of the revision, while neither $E_3$ nor $E_4$ is. Note that $E_1$ is also disabled at the first GRASPABLE clause by this second revision, but its classification is unchanged, since $E_1$ has a satisfactory proof through the second GRASPABLE clause (after the first revision is applied).

Now consider the example:

$E_5$: {HAS-BOTTOM, LIGHT-WEIGHT, HAS-CONCAVITY, UPWARD-CONCAVITY}, labeled as positive.

This example is particularly interesting because *no* possible revision to $\Gamma$ can change its classification as uncovered. That is, the fact that $E_5$ is not a CUP (according to $\Gamma$) is 'stable', in the sense that no changes to the open components of $\Gamma$ can invalidate the conclusion. Thus $\Gamma$ is not be repairable for $E_5$. The fact that patchable theories contain such stable information is fundamental and, as we will see below, plays a key role in theory patching.

## 2.2 Hardness

Despite the deceptive simplicity of the problem, theory patching is intractable for propositional theories for a wide range of revision functions. We say that a revision function $\rho$ for $\langle \Gamma, \Omega \rangle$ **allows deletion** if for all $c \in \Omega$, $\rho(c)$ contains a revision that deletes $c$. Clearly, if $\rho$ is unrestricted it allows deletion.

**Theorem 1 (Propositional hardness)** *For any fixed revision function $\rho$ which allows deletion, PPATCH($\Gamma, \Omega, \rho, \mathcal{E}$) is NP-complete.*

**Proof.** First, it is easy to see that PPATCH∈NP, since we can check in polynomial time if a given set of revisions yields a theory satisfying a set of labeled examples $\mathcal{E}$. We show that PPATCH is NP-hard by reducing the satisfiability problem (SAT) to propositional theory patching. Suppose we wish to find an assignment that makes an arbitrary CNF formula $S$ satisfiable, where

$$S = \bigwedge_{i=1}^{n} (\bigvee_{j=1}^{k_i} a_{ij}) \vee (\bigvee_{j=1}^{l_i} \neg b_{ij})$$

where for all $i, j, k$, and $l$, $a_{ij}$ and $b_{kl}$ are (not necessarily distinct) propositions.

Now let $\Gamma$ be the theory:

$$R \quad \leftarrow \quad D_1 \wedge D_2 \wedge \cdots \wedge D_n$$

$$
\begin{array}{llll}
D_1 & \leftarrow & a_{11} & \qquad D_1 & \leftarrow & \neg b_{11} \\
& \vdots & & & \vdots \\
D_1 & \leftarrow & a_{1k_1} & \qquad D_1 & \leftarrow & \neg b_{1l_1} \\
D_2 & \leftarrow & a_{21} & \qquad D_2 & \leftarrow & \neg b_{21} \\
& \vdots & & & \vdots \\
D_n & \leftarrow & a_{nk_n} & \qquad D_n & \leftarrow & \neg b_{nl_n} \\
\\
a_{11} & \leftarrow & \underline{A_{11}} & \qquad b_{11} & \leftarrow & \underline{B_{11}} \\
& \vdots & & & \vdots \\
a_{nk_n} & \leftarrow & \underline{A_{nk_n}} & \qquad b_{nl_n} & \leftarrow & \underline{B_{nl_n}}
\end{array}
$$

Note that the total number of theory components (propositions, clauses and literals) is O($N$), where $N$ is the number of literals in $S$. Let $\Omega$ be the set of components of $\Gamma$ consisting of the antecedent literals in the clauses $a_{ij} \leftarrow A_{ij}$ and $b_{ij} \leftarrow B_{ij}$. Let $\rho$ be a revision function permitting deletion of the antecedent literals in $\Omega$. Let $E$ be an example for $\Gamma$ which assigns all primitive propositions the value $F$.

Consider now the theory patching problem PPATCH($\Gamma, \Omega, \rho, \{\langle E, T\rangle\}$), where $E$ is assigned a positive label. Any set of permissible revisions to components in $\Omega$ amounts to a truth-value assignment to the propositions which satisfies $S$. For example, if the antecedent literal $A_{ij}$ is left alone, the proposition $a_{ij}$ will have the value false, whereas deleting the antecedent literal $A_{ij}$ from the clause $a_{ij} \leftarrow A_{ij}$ produces the clause $a_{ij} \leftarrow$, giving $a_{ij}$ the value true. Hence, if the theory can be repaired in polynomial time in $|\Gamma|$, a satisfying truth assignment can also be found in polynomial time in the size of $S$. □

For example, consider the CNF formula $S = (a \vee b) \wedge (\neg a \vee c) \wedge (\neg b)$ and its corresponding theory $\Gamma$ shown in Figure 3. Each revision to a primitive antecedent literal corresponds to deciding whether the head of the corresponding clause will be true (if the antecedent literal is deleted, giving a 'fact' clause, eg "$a \leftarrow T$") or false (if the antecedent literal, and

| | | | | | | |
|---|---|---|---|---|---|---|
| $R$ | $\leftarrow$ | D1, D2, D3 | | $R$ | $\leftarrow$ | D1, D2, D3 |
| D1 | $\leftarrow$ | a | | D1 | $\leftarrow$ | a |
| D1 | $\leftarrow$ | b | | D1 | $\leftarrow$ | b |
| D2 | $\leftarrow$ | ¬a | | D2 | $\leftarrow$ | ¬a |
| D2 | $\leftarrow$ | c | | D2 | $\leftarrow$ | c |
| D3 | $\leftarrow$ | ¬b | | D3 | $\leftarrow$ | ¬b |
| a | $\leftarrow$ | $\underline{A}$ | | a | $\leftarrow$ | |
| b | $\leftarrow$ | $\underline{B}$ | | b | $\leftarrow$ | B |
| c | $\leftarrow$ | $\underline{C}$ | | c | $\leftarrow$ | |

(a)                                              (b)

Figure 3: Theories for the SAT problem for the formula $S = (a \lor b) \land (\neg a \lor c) \land (\neg b)$. (a) The input theory according to the construction for Theorem 1. Open antecedent literals are underlined. (b) The theory corresponding to the satisfying assignment $a = $T, $b = $F, $c = $T.

hence the false value given by $E$, is retained). Hence, a set of revisions to $\Gamma$ that satisfies the labeled example $\langle \{\neg A, \neg B, \neg C\}, T \rangle$ corresponds to a variable assignment satisfying $S$. For example, the revision giving the theory in Figure 3(b) corresponds to the satisfying assignment $a = $T, $b = $F, $c = $T.

## 2.3 Reducing patching to the stability problem

The hardness result above shows that for nearly any choice of revision function, including the unrestricted case, the theory patching problem is intractable. In this section, we develop constructive proofs of the tractability of special cases of theory patching with an unrestricted revision function (as above, in the unrestricted case we drop the parameter $\rho$ when no ambiguity results). We do this by first reducing the patching problem to that of finding 'benign' revisions to a component. A benign revision to a component is one which allows the revised component to become closed without affecting the repairability of the theory. We then reduce the problem of finding benign revisions to that of determining the stability of examples in the theory, where a stable example is one whose classification is the same in all revised versions of the theory. Finally, we show that determining stability is tractable provided the theory satisfies a general monotonicity property with respect to revision.

We begin by defining the notion of benign revision precisely:

**Definition 4** *Given a patchable theory $\langle \Gamma, \Omega \rangle$, a revision $r$ on a component $c \in \Omega$ is **benign** for a labeled example $\langle E, l \rangle$ if $\langle r(c, \Gamma), \Omega \backslash \{c\} \rangle$ is repairable for $\langle E, l \rangle$. Similarly, $r$ is **benign** for a set $\mathcal{E}$ of labeled examples if $\langle r(c, \Gamma), \Omega \backslash \{c\} \rangle$ is repairable for $\mathcal{E}$.*

That is, a revision to a theory component is benign if the resulting theory is repairable for $\mathcal{E}$, even if the component becomes closed after the revision is performed (*i.e.*, the revision is not retractable). Note that if a theory is repairable for an example set $\mathcal{E}$, then for each open component there must exist a revision, possibly null, which is benign for $\mathcal{E}$.

**Definition 5** *The **propositional benign revision problem** PBENIGN$(\Gamma, \Omega, c, \mathcal{E})$ is:*

**Given:** *a patchable theory $\langle \Gamma, \Omega \rangle$, a component $c \in \Omega$, and a set of labeled examples $\mathcal{E}$,*

**Find:** *a revision $r$ on $c$ which is benign for $\mathcal{E}$, if such exists; otherwise return FAIL.*

The key to successful theory patching is finding benign revisions. This is seen via a simple reduction from PPATCH to PBENIGN, which is linear in $|\Omega|$.

**Theorem 2** *Given a patchable propositional theory $\langle \Gamma, \Omega \rangle$, and a training set of labeled examples $\mathcal{E}$, PPATCH is reducible to PBENIGN in time linear in $|\Omega|$.*

**Proof.** The following algorithm solves PPATCH, given PBENIGN as a subroutine. The idea is to perform a benign revision on each open component (recall that the null revision is also a revision).

**PPATCH**$(\Gamma, \Omega, \mathcal{E})$

1. $\Gamma' \leftarrow \Gamma$, $\Omega' \leftarrow \Omega$, $\mathcal{R} \leftarrow \emptyset$;

2. While $\Omega' \neq \emptyset$:

   (a) Choose $c \in \Omega'$;
   (b) $r \leftarrow$ **PBENIGN**$(\Gamma', \Omega', c, \mathcal{E})$;
   (c) If $r =$ FAIL, return FAIL;
   (d) $\Gamma' \leftarrow r(c, \Gamma')$;
   (e) $\Omega' \leftarrow \Omega' \backslash c$;
   (f) $\mathcal{R} \leftarrow \mathcal{R} \cup \{r\}$;

3. Return $\mathcal{R}$;

First note that the complexity of the algorithm is O($|\Omega| \cdot m$), where $m$ is the complexity of solving PBENIGN, since there is one call of PBENIGN for each component in $\Omega$.

Now we show that the algorithm solves PPATCH. First suppose that $\langle \Gamma, \Omega \rangle$ is repairable. We contend that output $\mathcal{R}$ is a set of revisions that produces a repaired theory from $\Gamma$, in other words, at the end of the algorithm, $\Gamma'$ classifies all examples in $\mathcal{E}$ correctly. To see this, consider the patchable theory $\langle \Gamma', \Omega' \rangle$ on each iteration of Step 2. Before the loop begins, $\langle \Gamma', \Omega' \rangle = \langle \Gamma, \Omega \rangle$ and so is repairable by assumption. After each execution of the body of the loop, we update $\langle \Gamma', \Omega' \rangle \leftarrow \langle r(c, \Gamma'), \Omega' \backslash \{c\} \rangle$ which is also repairable, since $r$ is benign. When the loop terminates, $\Omega' = \emptyset$, hence $\langle \Gamma', \emptyset \rangle$ is repairable (by the loop invariant). Thus $\Gamma'$ must classify all examples in $\mathcal{E}$ correctly, since no open components are available to repair.

Conversely, if $\langle \Gamma, \Omega \rangle$ is not repairable, there will be no benign repairs to any components, and hence the algorithm will return FAIL. □

We will now show how to reduce the problem of finding benign revisions to that of finding stable examples. First we will define the notion of example stability more precisely:

**Definition 6** *An example $E$ is **stably covered** by a patchable theory $\langle \Gamma, \Omega \rangle$ if, for every theory $\Gamma'$ obtainable from $\Gamma$, $\Gamma'(E) = T$. Similarly, $E$ is **stably uncovered** by $\langle \Gamma, \Omega \rangle$ if for every obtainable theory $\Gamma'$, $\Gamma'(E) = F$. $E$ is **stable** in $\langle \Gamma, \Omega \rangle$ if it is either stably covered or stably uncovered by $\langle \Gamma, \Omega \rangle$.*

We may now define the stability problem as follows.

**Definition 7** *The **propositional stability problem** $PSTABLE(\Gamma, \Omega, E)$ is:*

**Given:** *a patchable theory $\langle \Gamma, \Omega \rangle$ and an example $E$ for $\Gamma$,*

**Determine:** *if $E$ is stably covered or stably uncovered by $\langle \Gamma, \Omega \rangle$, returning $T$ if $E$ is stably covered, $F$ if $E$ is stably uncovered, and $U$ otherwise.*

Then we have that:

**Theorem 3** *Given a patchable propositional theory $\langle \Gamma, \Omega \rangle$, some distinguished component $c \in \Omega$, and a training set of labeled examples $\mathcal{E}$, PBENIGN is polynomially reducible to PSTABLE.*

**Proof.** We prove the theorem by giving an algorithm which constructs a benign repair. The algorithm works in two stages. First, we find two subsets of $\mathcal{E}$: a set of **obstructive** examples $O$, which are the examples for which $c$ must be disabled; and a set of **protected** examples $P$, which are the examples for which $c$ must not be disabled. Next, using the assumption that our revision function is unrestricted, we construct a revision to $c$ which disables $c$ for the examples in $O$ but not for those in $P$.

**PBENIGN**$(\Gamma, \Omega, c, \mathcal{E})$

1. $O \leftarrow \emptyset, P \leftarrow \emptyset$;

2. For each $\langle E, l \rangle \in \mathcal{E}$:

   (a) If **PSTABLE**$(\Gamma, \Omega \backslash \{c\}, E) = \neg l$, then $O \leftarrow O \cup \{E\}$;
   (b) If **PSTABLE**$(\Gamma \backslash \{c\}, \Omega \backslash \{c\}, E) = \neg l$, then $P \leftarrow P \cup \{E\}$;

3. If $O \cap P \neq \emptyset$, return FAIL;

4. Else:

    (a) If $c$ is an antecedent literal, then:

         i. If $P = \emptyset$, return the revision that deletes $c$;
        ii. Else, if $O = \emptyset$, return the null revision;
       iii. Else, return FAIL;

    (b) Else, return a revision with disabling set $D$ such that $O \subset D$ and $P \cap D = \emptyset$.

Now, it is easy to see that if $O$ and $P$ are not disjoint, then $\langle \Gamma, \Omega \rangle$ is not repairable for $\mathcal{E}$, since in such a case at least one example exists which $\langle \Gamma, \Omega \rangle$ stably classifies incorrectly. Conversely, if $\langle \Gamma, \Omega \rangle$ is not repairable for $\mathcal{E}$, there will be at least one example which is stably classified incorrectly, and that example will show up in both $O$ and $P$. Hence, the failure condition is satisfied.

Now suppose that $\langle \Gamma, \Omega \rangle$ is repairable for $\mathcal{E}$. We show that the patchable theory $\langle \hat{\Gamma}, \hat{\Omega} \rangle = \langle r(c, \Gamma), \Omega \backslash \{c\} \rangle$ is repairable, and hence $r$ is benign. Consider now the stability of examples in $\langle \hat{\Gamma}, \hat{\Omega} \rangle$:

- $\langle \hat{\Gamma}, \hat{\Omega} \rangle$ does not stably classify any example $E \in O$ incorrectly, since $c$ is disabled for $E$ in $\langle \hat{\Gamma}, \hat{\Omega} \rangle$, and $E \notin P$ hence $\textbf{PSTABLE}(\Gamma \backslash \{c\}, \Omega \backslash \{c\}, E) \neq \neg l$.

- $\langle \hat{\Gamma}, \hat{\Omega} \rangle$ does not stably classify any example $E \in P$ incorrectly, since $c$ is not disabled for $E$ in $\langle \hat{\Gamma}, \hat{\Omega} \rangle$, and $E \notin O$ hence $\textbf{PSTABLE}(\Gamma, \Omega \backslash \{c\}, E) \neq \neg l$.

- $\langle \hat{\Gamma}, \hat{\Omega} \rangle$ does not stably classify any example $E \in \mathcal{E} \backslash (P \cup O)$ incorrectly, since regardless of whether or not $c$ is disabled for $E$, both $\textbf{PSTABLE}(\Gamma \backslash \{c\}, \Omega \backslash \{c\}, E) \neq \neg l$, and $\textbf{PSTABLE}(\Gamma, \Omega \backslash \{c\}, E) \neq \neg l$ hold.

Since $\langle \hat{\Gamma}, \hat{\Omega} \rangle$ therefore does not stably classify any example incorrectly, it is repairable, and hence $r$ is benign. $\qquad \square$

We now close the circle by proving the equivalence of PPATCH and PSTABLE:

**Corollary 1** *PPATCH and PSTABLE can each be reduced to the other in polynomial time.*

**Proof.** That PPATCH can be reduced to PSTABLE in polynomial time follows directly from Theorems 2 and 3. In the other direction, PSTABLE$(\Gamma, \Omega, E)$ trivially reduces to PPATCH$(\Gamma, \Omega, \{\langle E, l \rangle\})$ where $l = \neg \Gamma(E)$, by noting that $\langle \Gamma, \Omega \rangle$ is repairable for $\{\langle E, l \rangle\}$ if and only if $E$ is not stable in $\langle \Gamma, \Omega \rangle$. $\qquad \square$

## 2.4 Determining example stability

Given the results of the previous section, we see that theory patching is tractable if and only if determining example stability is tractable. The question addressed in this section is: When is PSTABLE (and hence PPATCH) tractable? Intuitively, determining stability is intractable if revisions to a component may both permit new proofs and restrict old ones. Determining the stability of an example (hence its need to be repaired) may then require considering all possible combinations of components which could be revised, leading to a combinatorial explosion. We therefore consider here a monotonicity condition on theories, which requires that revising a component in the theory either permits proofs or restricts proofs, but not both. For example, in a propositional theory without negation, literals contribute solely towards preventing proofs, while clauses contribute solely towards facilitating them. We call a theory with such a monotonicity condition *parity-definite*, as defined below.

More precisely, we define the parity of each component in a theory as either *even*, *odd*, or *undefined*, as follows.

**Definition 8** *Given a propositional domain theory, define the* **parity** *of its root proposition to be* even. *Further, recursively define the* **parity** *of:*

- *a clause to be* even *(*odd*) if the parity of its head proposition is* odd *(*even*), and to be* undefined *otherwise;*

- *an antecedent literal to a clause to be* even *(*odd*) if the parity of its clause is* odd *(*even*), and to be* undefined *otherwise;*

- *an internal proposition to be* even *(*odd*) if every positive antecedent literal for that proposition is* even *(*odd*), and every negative antecedent literal containing that proposition is* odd *(*even*), and to be* undefined *otherwise.*

In general, odd components play the same role in a given theory as do clauses in negation-free theories, in that they only facilitate proofs of the root proposition in the theory, while even components play the role of antecedent literals, as they only restrict such proofs. Thus, revisions to odd components have the effect of specializing the theory, while revisions to even components have the effect of generalizing the theory. A theory that has only components with defined parity, therefore, will be monotonic in the sense discussed above. Hence we define:

**Definition 9** *A domain theory is* **parity-definite** *if every non-primitive component in the theory has a defined parity (even or odd).*

Note that we exclude primitive propositions from consideration, since only the parity of components that may in principle be revised is important. This fact is highlighted by the following extension of the definition of 'parity-definite' to patchable theories:

**Definition 10** *A patchable domain theory* $\langle \Gamma, \Omega \rangle$ *is* **parity-definite** *if every component in* $\Omega$ *has a defined parity.*

The restriction to parity-definite theories is syntactic, rather than semantic, since any theory may be reduced to an equivalent parity-definite theory. This is easily seen by reducing a theory to a DNF formula containing only primitive propositions; the 'flat' theory corresponding to such a formula is parity-definite. Thus, the restriction to parity definite theories takes on its primary significance when not all components of a theory are open.

We now have that theory patching of parity-definite patchable propositional theories is tractable:

**Theorem 4 (Propositional tractability)** *$PPATCH(\Gamma, \Omega, \mathcal{E})$ for parity-definite patchable theories $\langle \Gamma, \Omega \rangle$ is solvable in time polynomial in $|\Omega|$ and $|\mathcal{E}|$.*

This theorem follows immediately from Corollary 1 and the following lemma.

**Lemma 1** *If $\langle \Gamma, \Omega \rangle$ is parity-definite, $PSTABLE(\Gamma, \Omega, E)$ is solvable in time linear in $|\Gamma|$.*

**Proof.** Let $\Omega_E \subset \Omega$ be the set of the open even components of $\Gamma$, and $\Omega_O \subset \Omega$ be the set of the open odd components of $\Gamma$. Note that $\Omega = \Omega_E \cup \Omega_O$. Now let $\Gamma_{\text{gen}}$ be the theory which results from deleting the components in $\Omega_E$ from $\Gamma$, and let $\Gamma_{\text{spec}}$ be the theory which results from deleting the components in $\Omega_O$ from $\Gamma$. Since deleting components in $\Omega_O$ from $\Gamma$ can only make it harder to prove an example, an example is stably covered in $\Gamma$ if and only if it is covered in $\Gamma_{\text{spec}}$. Similarly, an example is stably uncovered in $\Gamma$ if and only if it is uncovered in $\Gamma_{\text{gen}}$. This gives the following linear algorithm for determining example stability:

**PSTABLE**$(\Gamma, \Omega, E)$

    1. $\Omega_E \leftarrow \{c \in \Omega \mid c \text{ even in } \Gamma\}$;

    2. $\Omega_O \leftarrow \{c \in \Omega \mid c \text{ odd in } \Gamma\}$;

    3. $\Gamma_{\text{gen}} \leftarrow \Gamma \backslash \Omega_E$;

    4. $\Gamma_{\text{spec}} \leftarrow \Gamma \backslash \Omega_O$;

    5. If $\Gamma_{\text{gen}}(E) = \Gamma_{\text{spec}}(E) = T$, then return $T$,

    6. Else if $\Gamma_{\text{gen}}(E) = \Gamma_{\text{spec}}(E) = F$, then return $F$,

    7. Else return $U$.                $\square$

For example, applying the algorithm to the cup theory above gives us the theories $\Gamma_{\text{gen}}$ and $\Gamma_{\text{spec}}$ shown in Figure 4. The example {HAS-BOTTOM, LIGHT-WEIGHT, HAS-CONCAVITY, UPWARD-CONCAVITY} is easily seen to be stably uncovered, as it is uncovered in both of these theories. On the other hand, the example {HAS-BOTTOM, LIGHT-WEIGHT, HAS-CONCAVITY, UPWARD-CONCAVITY, SMALL, DRY} is not stable, since it is covered in $\Gamma_{\text{gen}}$ but uncovered in $\Gamma_{\text{spec}}$.

| $\Gamma_{\text{gen}}$ | | |
|---|---|---|
| CUP | $\leftarrow$ | UPRIGHT $\wedge$ LIFTABLE $\wedge$ OPEN |
| UPRIGHT | $\leftarrow$ | HAS-BOTTOM |
| LIFTABLE | $\leftarrow$ | GRASPABLE $\wedge$ LIGHT-WEIGHT |
| OPEN | $\leftarrow$ | HAS-CONCAVITY $\wedge$ UPWARD-CONCAVITY |
| OPEN | $\leftarrow$ | HAS-STRAW |
| GRASPABLE | $\leftarrow$ | HAS-HANDLE |
| GRASPABLE | $\leftarrow$ | SMALL $\wedge$ DRY |
| $\Gamma_{\text{spec}}$ | | |
| CUP | $\leftarrow$ | UPRIGHT $\wedge$ LIFTABLE $\wedge$ OPEN |
| UPRIGHT | $\leftarrow$ | HAS-BOTTOM |
| LIFTABLE | $\leftarrow$ | GRASPABLE $\wedge$ LIGHT-WEIGHT |
| OPEN | $\leftarrow$ | HAS-CONCAVITY $\wedge$ UPWARD-CONCAVITY |
| OPEN | $\leftarrow$ | HAS-STRAW |
| GRASPABLE | $\leftarrow$ | SMALL $\wedge$ CERAMIC $\wedge$ DRY |

Figure 4: Theories generated in computing PSTABLE for the cup theory.

## 2.5 Soundness and Completeness

In this section we will show how the concepts of benign repair and example stability can be used in the analysis of soundness and completeness of theory patching algorithms. We consider here also algorithms that operate under restricted revision functions.

Any theory patching algorithm, *i.e.*, one that revises a given theory by patching individual open components of the theory, may be cast in the form of the following algorithm schema:

**Revise**$(\Gamma, \Omega, \rho, \mathcal{E})$

1. $\Gamma' \leftarrow \Gamma$;

2. While **termination-condition** does not hold:

   (a) Choose a component $c \in \Omega$;

   (b) Choose a revision $r \in \rho(c)$;

   (c) $\Gamma' \leftarrow r(\Gamma')$;

   (d) $\mathcal{R} \leftarrow \mathcal{R} \cup \{r\}$;

3. Either return $\mathcal{R}$ or FAIL.

A theory patching algorithm is **sound** if whenever the algorithm terminates it either returns a revision sequence $\mathcal{R}$ which repairs $\langle \Gamma, \Omega \rangle$ for $\mathcal{E}$, or it returns FAIL and $\langle \Gamma, \Omega \rangle$ is not repairable for $\mathcal{E}$. The soundness of a theory patching algorithm depends directly on whether certain of the revisions it performs are benign:

**Theorem 5 (Propositional soundness)** *A theory patching algorithm is sound if and only if (a) the last revision it performs to each component c is benign and (b) the algorithm returns FAIL only when no such benign revision exists.*

**Proof.** Suppose a given theory patching algorithm terminates and the last revision it performs to each component $c$ is benign. Then, by the argument given in the proof of Theorem 4, the final revised theory $\Gamma'$ is repairable for $\mathcal{E}$. On the other hand, if the algorithm returns FAIL, by assumption no benign repair exists for some component $c$, and so $\Gamma$ is not repairable. Hence the algorithm is sound.

Now suppose the given algorithm is sound and terminates for input $\langle \Gamma, \Omega \rangle$. Let $\Gamma'$ be the final revised theory in computed by the algorithm, and let $\langle \hat{\Gamma}, \hat{\Omega} \rangle$ be the patchable theory obtained prior to the last revision $r^*$ to $c$. Then $\langle r^*(c, \hat{\Gamma}), \hat{\Omega} \backslash \{c\} \rangle$ is repairable since $\Gamma'$ can be obtained from it by a sequence of permissible revisions. Thus $r^*$ is benign. □

Note that if the termination condition of a given theory patching algorithm is guaranteed to eventually obtain for every $\Gamma$, $\Omega$, and $\mathcal{E}$, the algorithm is also complete, in the sense that it is guaranteed to return a revision sequence $\mathcal{R}$ that repairs $\Gamma$ for $\mathcal{E}$, if one exists, and FAIL otherwise.

The difficulty of proving soundness and completeness of theory patching under a restricted revision function stems from the fact that in such cases it is difficult to find benign revisions. However, in the unrestricted case, we have a straightforward method for determining soundness and completeness. Recall the definitions of *obstructive* and *protected* from the proof of Theorem 3.

**Theorem 6** *For the case of unrestricted revisions, a theory patching algorithm which always terminates is sound and complete if and only if the last revision it performs to each component c disables c for every obstructive example at c and does not disable c for any protected example at c.*

**Proof.** This theorem follows directly from Theorem 5 and the proof of Theorem 3. □

For parity-definite theories this condition is easily checked. Two theory patching algorithms that operate on parity-definite theories with unrestricted revision are PTR (Koppel et al., 1994) and EITHER and its variants (Baffes & Mooney, 1994; Ourston & Mooney, 1994). The PTR algorithm uses a probabilistic heuristic to decide where to repair such that later revisions to a component are stronger than earlier revisions and the last revision to a component is guaranteed to be benign (Koppel et al., 1994). PTR can thus easily be shown to be sound and complete.

The EITHER algorithm patches theories without negation by first finding a subset of leaf antecedent literals (the **antecedent cover**) and a subset of leaf clauses (the **rule cover**) such that revisions to the two covers will suffice to repair the theory for the training set. Each component in each cover is revised once. The revision performed to a component ensures that (a) the component is disabled for all obstructive examples and that (b) the component is not disabled for any currently correctly classified example. Note, however, that (b) is weaker than the required condition of ensuring that the component is not disabled for any protected example, since (b) ignores incorrectly classified examples which are nevertheless protected. Thus, the heuristic argument in favor of the convergence of EITHER given by Ourston (1991) can only be formalized provided that condition (b) is equivalent to the required condition. This is the case only if all the components in the antecedent cover are revised first, and only then is the rule cover computed and its constituent rules revised. This constraint can easily be removed if revisions are chosen that satisfy the condition given in Theorem 6.

## 3. First-Order Theory Patching

### 3.1 Defining the problem

We now extend our treatment to cover the problem of theory patching for first-order theories. In this section, we consider first-order domain theories which are non-recursive, function-free, definite-clause theories with a single root predicate $R(x_1, \ldots, x_n)$, and negation-as-failure in clause antecedents. We assume as given a set of **fact predicates**, where a theory may contain **facts** each of which is a (possibly partial) instantiation of the parameters of a fact predicate in a bodiless clause. Facts with no free parameters are termed **ground facts**. For example, if $F(x, y)$ is a fact predicate, the fact $F(1, 2)$ is ground, while the fact $F(x, 4)$ is non-ground. A fact predicate cannot appear as the head of a clause with a non-empty body.

An **example** $E$ for a first-order theory $\Gamma$ with root $R(x_1, \ldots, x_n)$ is an assignment of the values $E^1, \ldots, E^n$ to the respective parameters $x_1, \ldots, x_n$. If $R(E^1, \ldots, E^n)$ is proved in $\Gamma$, $E$ is termed **covered** by $\Gamma$ (denoted $\Gamma(E) = T$), otherwise $E$ is termed **uncovered** by $\Gamma$ (denoted $\Gamma(E) = F$). A **labeled example** is a pair $\langle E, l \rangle$ consisting of an example $E$ and a truth assignment $l \in \{T, F\}$ for the instantiated root predicate $R(E^1, \ldots, E^n)$. If $l = T$, the labeled example is termed **positive**, otherwise it is termed **negative**.

The **components** of a first-order theory are its clauses, predicates, and the literals in the antecedent of each clause.

We now consider revisions to a first-order domain theory. As in the propositional case, we must first define what constitutes a revision to a component as opposed to a global revision to the theory.

- We regard deletion of a component to be a revision to that component. Thus we can revise a clause or an antecedent literal by simply deleting it from the theory. Note that we can think of deleting an antecedent literal as treating the literal as if it were always true. Hence, we naturally extend the definition to propositions by defining deletion of a non-primitive predicate $P(x_1, \ldots, x_n)$ to mean the addition of a clause asserting that $P(x_1, \ldots, x_n)$ is true for all instantiations of $x_1, \ldots, x_n$. We denote the theory resulting from the deletion of component $c$ from $\Gamma$ as $\Gamma \backslash \{c\}$.

- We may also revise a clause with head $P(x_1, \ldots, x_n)$ by adding a new positive or negative literal for new predicate $Q(y_1, \ldots, y_k)$ to the body of the clause, and adding a set of new clauses defining $Q$ to $\Gamma$. In order to keep the effects of the revision local to the clause under revision, we require that neither $Q$ nor any of the predicates appearing in its definition are non-fact predicates in $\Gamma$, and that $\{y_1, \ldots, y_k\} \subset \{x_1, \ldots, x_n\}$.

- Similarly, we may revise a predicate $P(x_1, \ldots, x_n)$ by adding a clause $C$ with head $P(x_1, \ldots, x_n)$ to $\Gamma$ along with a set of new clauses defining the literals in the body of $C$. In order to keep the effects of the revision local to the predicate under revision, we require that the body of no clause added to $\Gamma$ contains a non-fact predicate in $\Gamma$.

In the case of a revision to a clause with head $P(x_1, \ldots, x_n)$, the set of instantiations to the variable vector $\langle x_1, \ldots, x_n \rangle$ for which the new antecedent literal is false is termed the **disabling set** of the revision. Similarly, in the case of a revision to a predicate $P(x_1, \ldots, x_n)$, the set of instantiations to $\langle x_1, \ldots, x_n \rangle$ for which the body of the new clause is satisfied is termed the **disabling set** of the revision.

As in the propositional case, the defining property of a revision at a component is the set of instantiations for which the component is disabled. Note, however, that in the first order case, the disabling set of a revision to a component consists of instantiations for the variable vector appearing at *that* component. Thus a revision cannot be directly tailored to disable the component for individual examples, which are instantiations of the variable vector at the root.

Note that some of the more exotic revisions considered in the literature, such as changing the order of the variables in an antecedent literal, are not revisions to individual theory components as defined. They can, however, be constructed as combinations of revisions to individual components of the theory.

**Definition 11** *A* **patchable first-order theory** *is a pair $\langle \Gamma, \Omega \rangle$ consisting of a first-order domain theory $\Gamma$ and a set $\Omega$ of components of $\Gamma$. We call $\Omega$ the* **open components** *of $\langle \Gamma, \Omega \rangle$.*

A **revision function** $\rho$ and the notion of $\rho$-**obtainability** are defined as in the propositional case. We say that $\rho$ is **locally unrestricted**[3] if for each open predicate or clause component $c \in \Omega$ and set $\mathcal{I}$ of instantiations of $c$'s variable vector[4] there exists $r \in \rho(c)$ with disabling set $\mathcal{I}$.

Now we can define the first-order theory patching problem as in the propositional case.

**Definition 12** *The* **first-order theory patching problem** *$FPATCH(\Gamma, \Omega, \rho, \mathcal{E})$ is:*

**Given:** *a patchable first-order theory $\langle \Gamma, \Omega \rangle$ and a set $\mathcal{E}$ of labeled examples for $\Gamma$,*

**Find:** *a theory $\Gamma'$, $\rho$-obtainable from $\Gamma$, such that for all $E \in \mathcal{E}$, $\Gamma'(E) = l$, if such exists; otherwise return FAIL.*

---

3. We use the term '*locally* unrestricted' to suggest that only the effect of a revision on a local instantiation is determined directly; the effect of a revision on an example (an instantiation to the root's variable vector) depends globally on the structure of the theory.

4. The variable vector of a clause is the variable vector of the clause's head.

If such a $\rho$-obtainable theory $\Gamma'$ exists, we say that $\langle \Gamma, \Omega \rangle$ is **repairable** under $\rho$ for $\mathcal{E}$.

This definition is exactly parallel to that of the PPATCH patching problem for propositional theories, except that we consider first-order theories. FPATCH is considerably harder than PPATCH, however, as we will see in the next section. The restriction to parity-definite theories (defined exactly as in the propositional case) is not sufficient to ensure tractability, and the problem is intractable even in quite restricted settings.

## 3.2 Hardness

We consider in this section two restrictions on first-order domain theories, showing that even with these restrictions, the patching problem is quite difficult. A **completely bound** theory is one in which every variable appearing in an antecedent also appears in the head of that antecedent's clause. A **quasi-propositional** theory is a completely-bound theory in which every literal in the theory has the same variable vector as the root.

For example, the following theory is completely bound but not quasi-propositional:

```
R(x,y,z) :- Q(x,y) & S(y,z)
Q(x,y)   :- T(x)
Q(x,y)   :- T(y) & S(y,x)
```

The following theory is quasi-propositional:

```
R(x,y,z) :- Q(x,y,z) & S(x,y,z)
Q(x,y,z) :- T(x,y,z)
Q(x,y,z) :- T(x,y,z) & S(x,y,z)
```

However, the following theory is not quasi-propositional (since the order of the parameters changes):

```
R(x,y,z) :- Q(x,y,z) & S(x,y,z)
Q(x,y,z) :- T(y,x,z)
Q(x,y,z) :- T(x,y,z) & S(z,y,x)
```

**Theorem 7 (First-order hardness I)** *For any fixed revision function $\rho$ which allows deletion, FPATCH$(\Gamma, \Omega, \rho, \mathcal{E})$ is NP-hard, even if we consider only domain theories which:*

- *are negation free (hence parity-definite),*

- *are completely bound,*

- *are of bounded depth,*

- *contain no open clauses or propositions (only antecedent literals are open), and*

- *contain no non-ground facts.*

**Proof.** We prove the theorem by showing a reduction from the NP-complete problem MONOTONE-SAT (Garey & Johnson, 1979). Consider the monotone CNF expression $A$ over the set of variables $V = \{x_i\}$

$$A = \bigwedge_i (\bigvee_j p_{ij}) \wedge \bigwedge_k (\bigvee_l \neg q_{kl})$$

(where for all $i, j$ $p_{ij} \in V$ and for all $k, l$, $q_{kl} \in V$). We now construct a patchable theory $\langle \Gamma, \Omega \rangle$ and associated set of examples $\mathcal{E}$ such that the expression can be satisfied only if $\Gamma$ is repairable for $\mathcal{E}$, under any $\rho$ which allows deletion.

Let the root of $\Gamma$ be the predicate $R$, whose parameters are all the variables in $V$, plus a new variable $w$. Let $\Gamma$ be a theory as follows:

$$
\begin{aligned}
R(x_1, \ldots, x_n, w) \quad &\leftarrow \quad S(x_1, \ldots, x_n, w) \wedge T(x_1, \ldots, x_n, w) \\[4pt]
S(x_1, \ldots, x_n, w) \quad &\leftarrow \quad \mathsf{Zero}(w) \\
S(x_1, \ldots, x_n, w) \quad &\leftarrow \quad Q_1(x_1) \wedge \mathsf{Zero}(x_1) \\
S(x_1, \ldots, x_n, w) \quad &\leftarrow \quad Q_2(x_2) \wedge \mathsf{Zero}(x_2) \\
&\quad \vdots \\
S(x_1, \ldots, x_n, w) \quad &\leftarrow \quad Q_n(x_n) \wedge \mathsf{Zero}(x_n) \\[4pt]
T(x_1, \ldots, x_n, w) \quad &\leftarrow \quad \mathsf{One}(w) \\
T(x_1, \ldots, x_n, w) \quad &\leftarrow \quad Q_1(x_1) \wedge \cdots \wedge Q_n(x_n) \\[4pt]
Q_1(x) \quad &\leftarrow \quad \mathsf{One}(x) \\
Q_2(x) \quad &\leftarrow \quad \mathsf{One}(x) \\
&\quad \vdots \\
Q_n(x) \quad &\leftarrow \quad \mathsf{One}(x) \\[4pt]
\mathsf{Zero}(0). \\
\mathsf{One}(1).
\end{aligned}
$$

Note that $|\Gamma| = O(N)$ and that $\Gamma$ is negation free, is completely bound, has a depth of 3, and contains only ground facts.

Let the set of open components $\Omega$ be the set of all the antecedent literals to the clauses $Q_i(x) \leftarrow \mathsf{One}(x)$. Let $\rho$ be any revision function allowing deletion of the components in $\Omega$.

Let $\mathcal{E}$ be a set consisting of one labeled example for each conjunct in $A$, as follows. For each positive conjunct $C_i = (\bigvee_j p_{ij})$, we construct a negative labeled example for $R$, where the parameters $p_{ij}$ and $w$ are set to 0 and all other parameters take on the value 1. Similarly, for each negative conjunct $C_k = (\bigvee_l \neg q_{kl})$, we construct a positive labeled example for $R$, where parameters $q_{kl}$ are set to 0 and all other parameters (including $w$) are set to 1.

Now consider the patching problem FPATCH$(\Gamma, \Omega, \rho, \mathcal{E})$. Any permissible revision to $\Gamma$ consists of deleting some set of antecedent literals of the form $\mathsf{One}(x_i)$ from clauses $Q(x_i) \leftarrow \mathsf{One}(x_i)$. We now claim that any revision to $\Gamma$ which satisfies the examples determines a satisfying assignment $\sigma$ to $V$, where if the revision deletes antecedent literal $\mathsf{One}(x_i)$ from $\Gamma$, $\sigma$ assigns $x_i$ to TRUE, and otherwise assigns it to FALSE. Conversely, any such satisfying assignment determines a satisfying revision, hence if no satisfying revision exists, $A$ is unsatisfiable.

We now show that a revision satisfies an example $E_i \in \mathcal{E}$ if and only if the corresponding truth assignment for $V$ satisfies the example's corresponding conjunct $C_i$. This suffices to establish the theorem, since all examples in $\mathcal{E}$ must be satisfied simultaneously. Let $\Gamma'$ be the revised version of $\Gamma$, where $\sigma$ is the substitution corresponding to the revision.

Consider first an arbitrary positive conjunct in $A$, $C_p = \bigvee_j x_{i_j}$, whose corresponding negative example $E_p$ assigns each $x_{i_j}$ and $w$ to 0, and all other parameters to 1. The only way for the example not to be covered by $\Gamma'$ is for $T$ to be unprovable for $E$ in $\Gamma'$. This can only be the case if at least one of the literals $Q_{i_j}(0)$ is unprovable, and hence the antecedent literal to the corresponding clause in $\Gamma'$ is unrevised from $\Gamma$. This is the case exactly when $\sigma$ satisfies $C_p$.

Similarly, consider the negative conjunct $C_n = \bigvee_j \neg x_{i_j}$, with corresponding positive example $E$ assigning each $x_{i_j}$ to 0, and all other parameters (including $w$) to 1. The only way for $E$ to be covered by $\Gamma'$ is if $S$ is proved for $E$, which can only occur if some $Q_{i_j}(0)$ is proved, and hence the antecedent literal to the corresponding clause has been revised. This is the case exactly when $\sigma$ satisfies $C_n$. $\qquad\square$

Next we show that if we drop the restriction on non-ground facts (keeping all other restrictions), FPATCH is hard even for quasi-propositional theories.

**Theorem 8 (First-order hardness II)** *For any fixed revision function $\rho$ allowing deletion, $FPATCH(\Gamma, \Omega, \rho, E)$ is NP-hard, even if we only consider theories which:*

- *are negation-free (hence parity-definite),*

- *are of bounded depth,*

- *contain no open clauses or propositions (only antecedent literals are open), and*

- *are quasi-propositional (hence completely bound).*

**Proof.** Note that here we explicitly allow for non-ground facts. We prove the theorem, as above, by showing a reduction from MONOTONE-SAT (Garey & Johnson, 1979). Take as given a monotone CNF expression $A$ over the set of variables $V = \{x_i\}$

$$A = \bigwedge_i (\bigvee_j p_{ij}) \wedge \bigwedge_k (\bigvee_l \neg q_{kl})$$

(where for all $i, j$ $p_{ij} \in V$ and for all $k, l$, $q_{kl} \in V$). We now construct a patchable theory $\langle \Gamma, \Omega \rangle$ and associated set of examples $\mathcal{E}$ such that $A$ can be satisfied only if $\Gamma$ is repairable for $\mathcal{E}$, under any revision function $\rho$ which allows deletion.

Let the root of $\Gamma$ be the predicate $R$, whose parameters are all the variables in $V$, plus a new variable $w$. Let $\Gamma$ be a theory as follows:

$$R(x_1, \ldots, x_n, w) \;\leftarrow\; S(x_1, \ldots, x_n, w) \wedge T(x_1, \ldots, x_n, w)$$

$$S(x_1, \ldots, x_n, w) \;\leftarrow\; \mathsf{Zero}_w(x_1, \ldots, x_n, w)$$
$$S(x_1, \ldots, x_n, w) \;\leftarrow\; Q_1(x_1, \ldots, x_n, w) \wedge \mathsf{Zero}_1(x_1, \ldots, x_n, w)$$
$$S(x_1, \ldots, x_n, w) \;\leftarrow\; Q_2(x_1, \ldots, x_n, w) \wedge \mathsf{Zero}_2(x_1, \ldots, x_n, w)$$
$$\vdots$$
$$S(x_1, \ldots, x_n, w) \;\leftarrow\; Q_n(x_1, \ldots, x_n, w) \wedge \mathsf{Zero}_n(x_1, \ldots, x_n, w)$$

$$T(x_1, \ldots, x_n, w) \;\leftarrow\; \mathsf{One}_w(x_1, \ldots, x_n, w)$$
$$T(x_1, \ldots, x_n, w) \;\leftarrow\; Q_1(x_1, \ldots, x_n, w) \wedge \cdots \wedge Q_n(x_1, \ldots, x_n, w)$$

$$Q_1(x_1, \ldots, x_n, w) \;\leftarrow\; \mathsf{One}_1(x_1, \ldots, x_n, w)$$
$$Q_2(x_1, \ldots, x_n, w) \;\leftarrow\; \mathsf{One}_2(x_1, \ldots, x_n, w)$$
$$\vdots$$
$$Q_n(x_1, \ldots, x_n, w) \;\leftarrow\; \mathsf{One}_n(x_1, \ldots, x_n, w)$$

$\mathsf{Zero}_1(0, x_2, x_3, \ldots, x_n, w). \qquad \mathsf{One}_1(1, x_2, x_3, \ldots, x_n, w).$

$\mathsf{Zero}_2(x_1, 0, x_3, \ldots, x_n, w). \qquad \mathsf{One}_2(x_1, 1, x_3, \ldots, x_n, w).$

$\vdots \qquad\qquad\qquad\qquad\qquad\qquad \vdots$

$\mathsf{Zero}_n(x_1 \ldots, x_{n-1}, 0, w). \qquad \mathsf{One}_n(x_1 \ldots, x_{n-1}, 1, w).$

$\mathsf{Zero}_w(x_1, \ldots, x_{n-1}, x_n, 0). \qquad \mathsf{One}_w(x_1, \ldots, x_{n-1}, x_n, 1).$

Note that $|\Gamma| = O(N)$ where $N$ is the number of literals in $A$, and that $\Gamma$ is negation free, is quasi-propositional (hence completely-bound), and has a depth of 3.

Let the set of open components $\Omega$ be the set of all the antecedent literals to the clauses $Q_i(x_1, \ldots, x_n, w) \leftarrow \mathsf{One}_i(x_1, \ldots, x_n, w)$. Let $\rho$ be any revision function allowing deletion of the components in $\Omega$.

Let $\mathcal{E}$ be a set consisting of one labeled example for each conjunct in $A$, as follows. For each positive conjunct $C_i = (\bigvee_j p_{ij})$, we construct a negative labeled example for $R$, where the parameters $p_{ij}$ and $w$ are set to 0 and all other parameters take on the value 1. Similarly, for each negative conjunct $C_k = (\bigvee_l \neg q_{kl})$, we construct a positive labeled example for $R$, where parameters $q_{kl}$ are set to 0 and all other parameters (including $w$) are set to 1.

An argument isomorphic to that given in the proof of Theorem 7 gives that $A$ can be satisfied only if $\mathrm{FPATCH}(\Gamma, \Omega, \rho, \mathcal{E})$ is solvable, $i.e.$, $\langle \Gamma, \Omega \rangle$ is repairable for $\mathcal{E}$. $\qquad\square$

Note that the constructions given above can be modified easily to replace the restriction of bounded theory depth with a restriction of bounded theory width.

### 3.3 Tractability

In the previous section we showed that FPATCH (for restricted and unrestricted revision functions) is intractable even if we severely restrict the class of theories considered such that, in addition to being negation-free, of bounded depth, and with all open components antecedent literals, theories either (a) contain no non-ground facts **or** (b) are quasi-propositional. We now show that FPATCH with locally unrestricted revisions is tractable for parity-definite theories for which **both** (a) and (b) hold, even without the other restrictions.

As in the propositional case, we drop the $\rho$ when talking about FPATCH under locally unrestricted revision functions.

**Theorem 9 (First-order tractability)** *For locally unrestricted revision functions, $FPATCH(\Gamma, \Omega, \mathcal{E})$ for parity-definite quasi-propositional theories with no non-ground facts is solvable in time polynomial in $|\Omega|$ and $|\mathcal{E}|$.*

This theorem is actually quite weak, since the restrictions allow a straightforward reduction to the propositional case.

**Proof.** The reduction to PPATCH is as follows.

- Let $\hat{\Gamma}$ be the propositional theory obtained from $\Gamma$ by replacing each literal $P(x_1, \cdots, x_n)$ by the proposition $\hat{p}$.

- Let $\hat{\Omega}$ consist of the components of $\hat{\Gamma}$ which correspond to the components in $\Omega$.

- Let the primitive propositions of $\hat{\Gamma}$ be those which correspond to literals for fact predicates in $\Gamma$.

- For any labeled example $\langle E, l \rangle$, let the corresponding propositional labeled example, $\langle \hat{E}, \hat{l} \rangle$, be such that $\hat{E}$ is a truth assignment to the primitive propositions of $\hat{\Gamma}$ such that for each $\hat{p}$, $\hat{E}(\hat{p}) = T$ if and only if $P(E^1, \cdots, E^n)$ is a ground fact in $\Gamma$, and $\hat{l} = l$. Also, for any set of labeled examples $\mathcal{E}$, let $\hat{\mathcal{E}}$ be the corresponding set of propositional labeled examples.

- For a propositional repair $\hat{r}$ on propositional component $\hat{c} \in \hat{\Gamma}$ with disabling set $\hat{D}$, the corresponding first-order revision $r$ on first-order component $c \in \Gamma$ is defined as the revision with disabling set $D$.

It then follows that any sequence of revisions $\hat{\mathcal{R}} = \{\hat{r}_i\}$ that solves PPATCH$(\hat{\Gamma}, \hat{\Omega}, \hat{\mathcal{E}})$ yields a corresponding sequence of revisions $\mathcal{R} = \{r_i\}$ which solves FPATCH$(\Gamma, \Omega, \mathcal{E})$. Moreover, if no positive solution exists for PPATCH$(\hat{\Gamma}, \hat{\Omega}, \hat{\mathcal{E}})$ then none exists for FPATCH$(\Gamma, \Omega, \mathcal{E})$.
$\square$

### 3.4 Discussion

How do our results in first-order relate to the propositional case? In the propositional case, we saw that sufficient conditions for tractability of theory patching were (a) that the theory be parity-definite, and (b) that the revision function be unrestricted. In the first-order case, however, we see that local unrestrictedness is a much weaker condition than having unrestricted revisions in the propositional case. This is because a revision of a component in first-order only affects the instantiations of the component's variable vector and does not directly affect examples, *i.e.*, instantiations of the root's variable vector. In fact, the additional conditions for tractability imposed above on the structure of the theory implicitly constitute a stronger unrestrictedness condition, which can be made explicit as follows.

A revision function $\rho$ on a theory $\Gamma$ is **truly unrestricted** if $\rho$ and $\Gamma$ are such that given an open clause or predicate component in $\Gamma$ and any two disjoint sets of *examples A* and *B*, a revision exists which can disable the component for every example in $A$ without disabling the component for any example in $B$. This condition fails if, for instance, two different examples agree on all the variables appearing at the component, or if a predicate appears more than once in the theory each time with a different variable vector, so that the predicate cannot be revised for one example without affecting other examples. Having truly unrestricted revisions makes the problem tractable for parity-definite theories, since the effect of a revision may be evaluated locally at a single component, without requiring consideration of other possible revisions at other components.

## 4. Conclusions

In this paper we have developed some necessary and some sufficient conditions for the soundness and tractability of theory patching. We have thus addressed the problem of theory revision in general, without restricting our focus to any particular algorithm. The central notion in theory patching is that of the benign revision, *i.e.*, that any non-retractable revision must ensure the repairability of the resulting theory. Since benign revisions must rule out stably misclassified examples, the key to theory patching is determining example stability. In fact, the two problems, theory patching and determining stability, are polynomially equivalent. In general, the patching problem is hard, but we have found that theory patching is tractable if the input theory is parity-definite (*i.e.*, monotonic with respect to revision) and revisions are truly unrestricted (*i.e.*, any open component can be disabled for an arbitrary set of input examples). Moreover, we have shown how the soundness of a theory revision method can be checked by verifying that certain of the revisions it performs are benign.

We have shown how to find *some* hypothesis obtainable from the input theory which is consistent with the given training examples, but we have not considered here the quality of the hypothesis obtained, in terms of convergence to some target concept (say, in the PAC learning framework (Valiant, 1984)). It is worth noting that the very condition of truly unrestricted revisions, which we have shown is useful for ensuring that patching is tractable, leads to difficulty in learnability, since the hypothesis space is then very large. In fact, in the truly unrestricted case, the VC-dimension of the hypothesis space may be as large as the total number of possible examples, since the set of truly unrestricted revisions at a component shatters arbitrary sets of examples at that component. Thus the question of what conditions are required for convergence of theory patching to a target theory remains open.

## Acknowledgements

We would like to thank William Cohen and the anonymous reviewers for their comments on previous drafts of this paper which helped improve it greatly. We would also like to thank Rina Schwartz for her comments on an earlier draft of this paper.